\documentclass[runningheads]{llncs}

\usepackage{eccv}
\usepackage[numbers,sort&compress]{natbib}
\usepackage{listings}
\usepackage{xcolor}

\usepackage{amssymb}      
\usepackage[table]{xcolor} 
\usepackage{colortbl}      
\usepackage{booktabs}      
\usepackage{multirow}
\usepackage{booktabs}
\usepackage{eccvabbrv}
\usepackage{tcolorbox}  
\usepackage{graphicx}
\usepackage{booktabs}
\usepackage{makecell}
\usepackage[accsupp]{axessibility}  

\author{Yupeng Gao\inst{1}}

\authorrunning{Y. Gao}

\institute{
	University of Electronic Science and Technology of China, Chengdu, China\\
	\email{202411081510@std.uestc.edu.cn}
}

\begin{document}

\title{UAV as Urban Construction Change Monitor: A New Benchmark and Change Captioning Model}
\titlerunning{Change Captioning for Urban Construction Monitoring}

\author{
	Yupeng Gao\inst{1} \and
	Tianyu Li\inst{1} \and
	Guoqing Wang\inst{1} \and
	Yang Yang\inst{1}
}

\authorrunning{Y. Gao et al.}

\institute{
	University of Electronic Science and Technology of China, Chengdu, China\\
	\email{
		202411081510@std.uestc.edu.cn,
		cosmos.yu@hotmail.com,
		gqwang0420@uestc.edu.cn,
		yang.yang@uestc.edu.cn
	}
}

\maketitle

\begin{abstract}
Remote Sensing Image Change Captioning (RSICC) aims to generate 
spatially grounded natural language descriptions of scene evolution 
from bi-temporal imagery, moving beyond binary change masks toward 
semantic-level understanding. However, existing methods rely on 
implicit feature differencing without explicitly modeling structured 
change semantics, and struggle to reconcile the conflicting 
representation demands of change detection and caption generation. 
In addition, current benchmarks provide limited coverage of 
high-resolution urban construction scenarios. To address these challenges, we propose PTNet, a prototype-guided 
task-adaptive framework for joint change captioning and detection. 
PTNet explicitly models structured change semantics through a 
learnable prototype bank that guides cross-temporal interaction, 
disentangles task-specific representations via multi-head gating, 
and injects detection-derived spatial priors into caption generation, 
enabling coherent semantic correspondence while preserving 
fine-grained spatial sensitivity. Furthermore, we construct UCCD, a large-scale UAV-based benchmark 
comprising 9,000 high-resolution image pairs and 45,000 annotated 
sentences for urban construction monitoring. Extensive experiments 
on UCCD and WHU-CDC demonstrate that PTNet consistently outperforms 
existing methods. The dataset and source code are publicly available 
at \url{https://github.com/G124556/ptnet}.
	
\end{abstract}

\section{Introduction}
The continuous advancement of remote sensing technologies has driven rapid 
progress in multi-temporal analysis tasks such as change detection and change 
captioning~\cite{hoxha2022change,zhu2025change3d,li2025cd4c}. 
Unlike traditional change detection that only localizes changed regions, 
Remote Sensing Image Change Captioning (RSICC) advances toward semantic-level 
understanding by generating natural language descriptions of scene evolution, 
providing interpretable information for land-use auditing, urban monitoring, 
and emergency response~\cite{liu2024pixel}, and enabling seamless integration 
into higher-level reasoning pipelines~\cite{peng2024change}.

\begin{figure}[t]
	\centering
	\includegraphics[width=\linewidth]{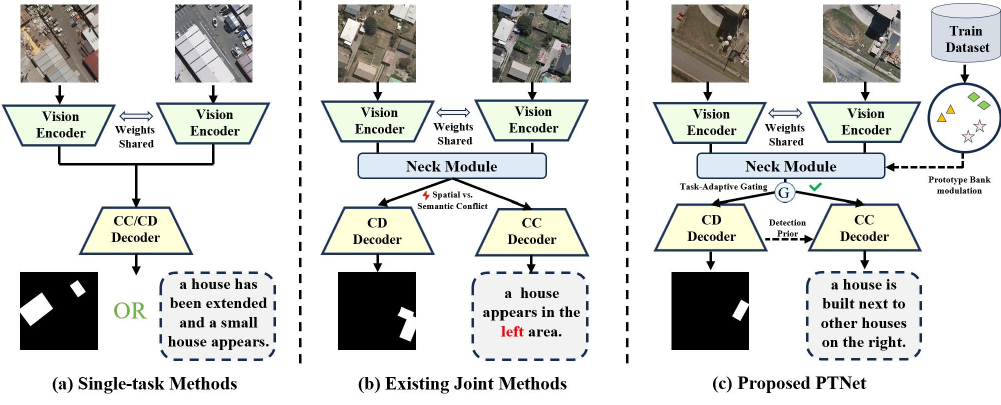}
	\caption{(a) single-task methods that produce either a change mask or a 
		caption, (b) existing joint methods that suffer from feature conflicts and 
		inaccurate descriptions, and (c) the proposed PTNet, which introduces 
		prototype-guided semantic modeling and task-adaptive feature decoupling 
		for accurate and spatially faithful change captioning.}
	\label{fig:motivation}
\end{figure}

Existing RSICC methods follow an encoder--decoder paradigm to model cross-temporal 
visual differences. While subsequent advances incorporating Transformers~\cite{vaswani2017attention,dosovitskiy2021vit}, 
state space models~\cite{meng2025rsic}, diffusion models~\cite{sun2025mask}, 
and large language models~\cite{zhang2024rsllava,leon2026describing} have 
progressively enhanced semantic alignment, and joint detection-captioning 
frameworks~\cite{vandenhende2021mtl,shi2024multitask,li2024detection} leverage 
spatial cues to improve caption quality, two fundamental limitations persist.
First, most approaches rely on implicit feature differencing without explicitly 
modeling structured change-type semantics, limiting accurate temporal 
correspondences in complex scenes~\cite{liu2024changeagent}. 
Second, change detection and captioning impose inherently incompatible 
representation demands---spatial precision versus semantic abstraction---and 
simple shared-feature or parallel-branch designs fail to resolve this 
granularity mismatch~\cite{sun2026scnet}, as illustrated in 
Fig.~\ref{fig:motivation}.

Beyond methodology, existing benchmarks provide insufficient coverage of urban 
construction scenarios, constrained in annotation scale and scene 
diversity~\cite{liu2025remote}. Most focus on natural disaster or agricultural 
changes, with sparse samples capturing fine-grained urban dynamics such as 
illegal building expansion and incremental land-use transitions, restricting 
systematic study in realistic urban environments.

To address these challenges, we propose \textbf{PTNet} (Prototype-Guided 
Task-Adaptive Network), a unified framework for joint change detection and 
captioning, where captioning serves as the primary objective and detection 
provides spatial grounding. PTNet models structured change semantics via 
learnable change-type prototypes for cross-temporal alignment, and employs 
task-adaptive gating to produce sub-task-aware representations. A lightweight 
mask encoder injects detection-derived spatial priors into caption generation, 
while a CLIP-based contrastive objective~\cite{radford2021clip,oord2018cpc} 
enforces cross-modal consistency. In addition, we construct UCCD, a 
large-scale UAV-based urban change captioning benchmark.
The main contributions are as follows:

\begin{enumerate}
	
	\item We introduce \textbf{PG-CAI} as a core semantic modeling component 
	of PTNet, which clusters training-set change features into learnable 
	change-type prototypes to guide cross-temporal attention and enable 
	explicit semantic-level temporal correspondences and structured semantic 
	priors for both sub-tasks.
	
	\item We develop \textbf{TAMG} module to learn task-adaptive feature 
	representations at the attention-head granularity, enabling detection and 
	captioning to maintain appropriate spatial and semantic sensitivities 
	within a unified framework, alleviating the conflict from 
	their incompatible feature requirements.
	
	\item A lightweight mask encoder is incorporated into PTNet to 
	inject detection-derived spatial priors into caption generation, while a 
	CLIP-based contrastive objective enforces cross-modal semantic consistency.
	
	\item We construct \textbf{UCCD}, a large-scale benchmark comprising 
	9{,}000 high-resolution low-altitude UAV image pairs and 45{,}000 
	annotated sentences for fine-grained urban construction change 
	understanding, providing both detection annotations and detailed semantic 
	descriptions. To our knowledge, UCCD is the first low-altitude UAV-based 
	benchmark specifically designed for joint change detection and captioning 
	in urban construction environments.
	
\end{enumerate}

\section{Related Work}
Remote sensing change detection has evolved from CNN-based 
approaches~\cite{fang2022snunet,he2016resnet,daudt2018fully,chen2020spatial} 
to Transformer-based architectures~\cite{liu2021swin,noman2024scratchformer}, 
with methods like BIT~\cite{chen2021bit} and 
ChangeFormer~\cite{bandara2022changeformer} improving localization accuracy 
via self-attention. However, their outputs remain limited to binary change 
masks. To address this, RSICC has emerged as a cross-modal task generating 
natural language descriptions of semantic 
changes~\cite{li2024intertemporal}.

Early RSICC methods established the encoder--decoder paradigm for 
cross-temporal visual understanding. RSICCformer~\cite{liu2022levir} 
introduced dual-branch Transformers to model bi-temporal feature 
interactions, while Chg2Cap~\cite{chang2023chg2cap} incorporated attentive 
difference modules to highlight changed regions. Subsequently, 
PromptCC~\cite{liu2023promptcc} introduced a prompt-based decoupling 
paradigm to disentangle change-relevant semantics; 
RSCaMa~\cite{liu2024rscama} leveraged state space models to capture 
long-range temporal dependencies; and MADiffCC~\cite{yang2024madiffcc} 
employed diffusion models to enrich caption diversity. More recently, 
Semantic-CC~\cite{liu2024semanticcc} and KCFI~\cite{yang2025kcfi} 
incorporated large language models to produce richer semantic descriptions. 
Despite these advances, all of the above methods rely on implicit feature 
differencing without explicitly modeling structured change-type semantics, 
and cannot leverage the complementary spatial supervision from change 
detection.

With the advancement of 
MLLMs~\cite{liu2023llava,li2023blip2,wang2022rvsa,lobry2020rsvqa,
	kuckreja2024geochat,liu2023earthvqa}, ChangeChat~\cite{deng2025changechat}, 
CDChat~\cite{noman2024cdchat}, and BTCChat~\cite{li2025btcchat} have 
progressively enriched semantic depth and temporal modeling in change 
understanding, demonstrating that integrating LLM reasoning with 
change-aware perception is a promising direction---motivating our 
LLM-based captioning branch with contrastive cross-modal alignment.

Beyond pure captioning, another active direction jointly models detection 
and captioning to leverage region-level supervision for more spatially 
faithful language generation~\cite{vandenhende2021mtl}. Most existing 
approaches adopt a shared backbone with dual task-specific heads, yet 
several design limitations remain.
Shi~\emph{et al.}~\cite{shi2024multitask} employs largely shared 
representations without explicitly accounting for task-specific feature 
sensitivities; Wang~\emph{et al.}~\cite{wang2024changeminds} introduces detection 
cues but does not effectively propagate spatial priors into the captioning 
branch; and Liu~\emph{et al.} and Li~\emph{et al.}~\cite{liu2024changeagent,
	li2024detection} do not explicitly model structured change types nor 
provide fine-grained task-aware representation control. In contrast, PTNet 
explicitly decouples task-specific representations via TAMG and injects 
detection spatial priors through mask-guided encoding, enabling the two 
tasks to mutually reinforce each other.

\section{Methodology}

\subsection{Overall Framework}

\begin{figure*}[t]
	\centering
	\includegraphics[width=0.7\textwidth]{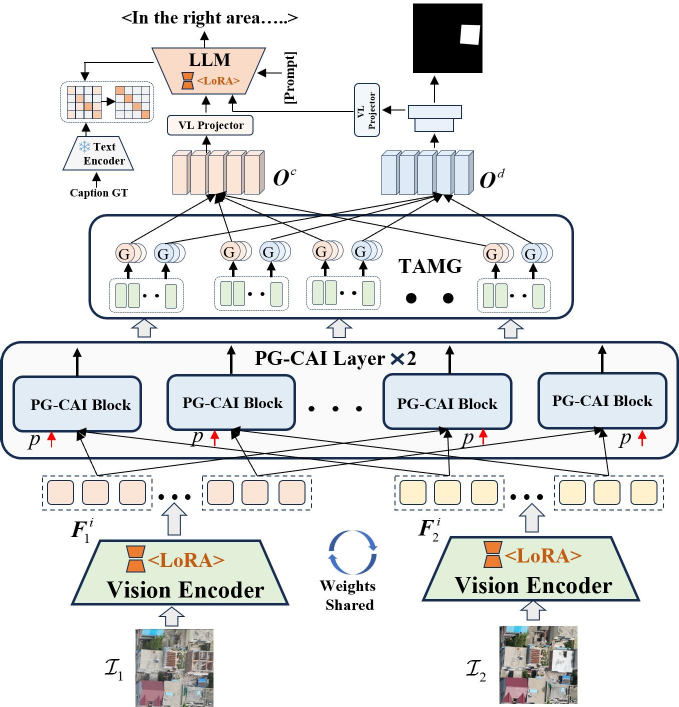}
	\caption{Overall architecture of the proposed PTNet.}
	\label{fig:framework}
\end{figure*}

 The overall architecture of PTNet, illustrated in Fig.~\ref{fig:framework}. Given a bi-temporal remote sensing image pair $(\mathcal{I}_1, \mathcal{I}_2) \in \mathbb{R}^{H \times W \times 3}$, 
our primary objective is to generate accurate and spatially-aware natural language 
descriptions $\mathbf{T} = [w_1, \ldots, w_L]$, while auxiliarily producing a change 
detection mask $\mathbf{M} \in \{0,1\}^{H \times W}$ as a spatial localization.

We employ a pre-trained CLIP Vision Encoder (ViT-L/14) as the 
shared backbone, fine-tuned via LoRA~\cite{hu2022lora}, to extract hierarchical feature representations from the 6th, 12th, 18th, 
and 24th transformer layers for each input image:
\begin{equation}
	\mathbf{F}_j^i = \Phi_v^{i}(\mathcal{I}_j), \quad j \in \{1, 2\},\ i \in \{1,2,3,4\},
\end{equation}
where $j$ denotes the temporal phase, $i$ denotes the hierarchy level, and 
$\mathbf{F}_j^i \in \mathbb{R}^{N \times D}$. The Prototype-Guided Change-Aware Interaction 
(PG-CAI) module maintains learnable change-type prototypes $\mathbf{P} \in \mathbb{R}^{K 
	\times N \times D}$ to guide prototype-modulated bidirectional cross-temporal attention, 
producing change-aware bi-temporal features $\{\mathbf{G}_1^i, \mathbf{G}_2^i\}$. The 
Task-Adaptive Multi-head Gating (TAMG) module then learns differentiated gating weights 
at the attention-head granularity to separately generate detection-oriented features 
$\mathbf{O}^d$ and captioning-oriented features $\mathbf{O}^c$ via cross-level adaptive 
fusion. In the task-specific decoding stage, an FPN~\cite{lin2017fpn} detection branch decodes 
$\mathbf{O}^d$ into a change probability map $\hat{\mathbf{M}}$, while a captioning 
branch built on Qwen2-1.5B-Instruct, fine-tuned via LoRA, encodes fine-grained spatial 
features extracted from the FPN detection branch into compact detection tokens, which are 
concatenated with $\mathbf{O}^c$ to inject spatial localization priors into caption 
generation. Finally, a vision-language semantic alignment loss based on InfoNCE pulls 
the captioning visual representations toward their corresponding CLIP text embeddings, 
ensuring consistency with the language semantic space.

\subsection{Prototype-Guided Change-Aware Interaction}

%

\subsubsection{Prototype Initialization}
\begin{figure}[t]
	\centering
	\includegraphics[width=\columnwidth]{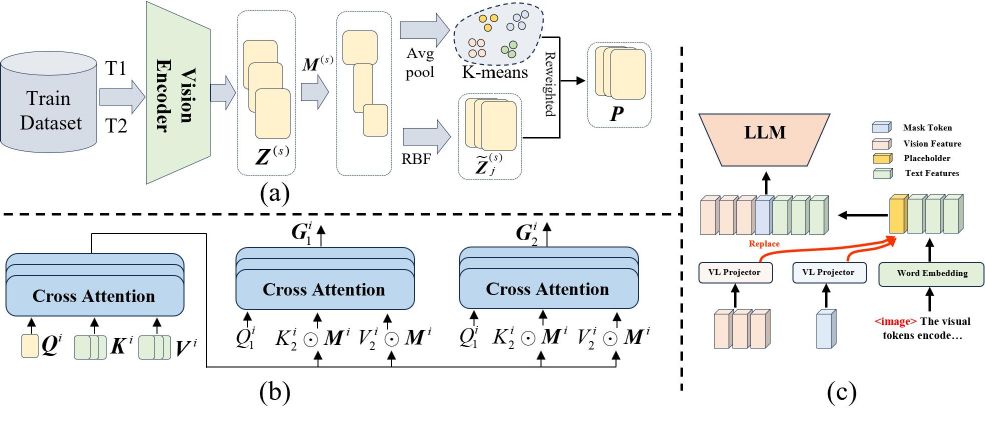}
	\caption{ (a)~\textbf{Prototype bank construction}: training-set difference features 
		are clustered via K-means and spatially recovered via RBF interpolation 
		to form the learnable prototype bank $\mathbf{P} \in \mathbb{R}^{K \times N \times D}$. 
		(b)~\textbf{PG-CAI Block}: $\mathbf{P}$ modulates bidirectional 
		cross-attention between $\{\mathbf{F}_1^i, \mathbf{F}_2^i\}$, producing 
		change-aware features $\{\mathbf{G}_1^i, \mathbf{G}_2^i\}$. 
		(c)~\textbf{Change Captioning Decoder}: change-aware features are projected 
		into the LLM token space and concatenated with detection tokens 
		for caption generation.}
	
	\label{fig:pgcai}
\end{figure}

As depicted in Fig.~\ref{fig:pgcai}(a), we initialize 
$\mathbf{P} \in \mathbb{R}^{K \times N \times D}$ offline using the training set, 
where $K$ is the number of prototype clusters, $N$ is the number of spatial 
tokens, and $D$ is the feature dimension. For each training sample $s$, we compute 
difference features using the second-level hierarchical features $\mathbf{F}_j^2$ 
extracted from the 12th transformer layer of the CLIP Vision Encoder, which captures 
mid-level semantic structure while retaining sufficient spatial resolution 
for change localization. A compact representation $\mathbf{z}^{(s)} \in \mathbb{R}^D$ 
is then obtained by mask-guided average pooling over changed pixels, or global 
average pooling for unchanged samples:
\begin{equation}
	\mathbf{Z}^{(s)} = |\mathbf{F}_1^{2,(s)} - \mathbf{F}_2^{2,(s)}| 
	\in \mathbb{R}^{N \times D}, \qquad
	\mathbf{z}^{(s)} = 
	\begin{cases}
		\mathrm{AvgPool}(\{\mathbf{Z}_b^{(s)}\}_{b \in \Omega^{(s)}}) 
		& \text{if } |\Omega^{(s)}| > 0 \\[4pt]
		\tfrac{1}{N}\sum_{b=1}^{N}\mathbf{Z}_b^{(s)} 
		& \text{otherwise}
	\end{cases}
\end{equation}
where $\Omega^{(s)} = \{b \mid \mathbf{M}_b^{(s)} = 1\}$ denotes the 
set of changed pixel indices. K-means clustering on 
$\{\mathbf{z}^{(s)}\}_{s=1}^S$ yields $K$ cluster centers 
$\{\mathbf{c}_k^0\}_{k=1}^K$, naturally covering one ``no-change'' 
pattern and $K{-}1$ distinct change types.

Since the pooling operation discards spatial structure, we recover it 
by re-expanding each aggregated vector into a dense spatial map 
$\tilde{\mathbf{Z}}^{(s)} \in \mathbb{R}^{N \times D}$ via Radial 
Basis Function (RBF) interpolation~\cite{11154010}:
\begin{equation}
	\tilde{\mathbf{Z}}_v^{(s)} = 
	\sum_{l \in \Omega^{(s)}} w_{vl}\,\mathbf{Z}_l^{(s)}, 
	\quad 
	w_{vl} \propto \exp\!\left(
	-\frac{\|\mathbf{p}_v - \mathbf{p}_l\|^2}{2\sigma^2}\right),
\end{equation}
where $\mathbf{p}_v, \mathbf{p}_l \in \mathbb{R}^2$ denote the 2D 
spatial coordinates of positions $v$ and $l$ respectively, and 
$\sigma$ is the Gaussian bandwidth controlling interpolation smoothness. 
For unchanged samples where $\Omega^{(s)} = \emptyset$, the 
interpolation reduces to uniform spatial replication. Finally, prototypes are initialized by aggregating spatially recovered 
features weighted by soft cluster assignments:
\begin{equation}
	\alpha_k^{(s)} = \frac{
		\exp(-\|\mathbf{z}^{(s)} - \mathbf{c}_k^0\|^2 / \tau)
	}{
		\sum_{k'=1}^K \exp(-\|\mathbf{z}^{(s)} - 
		\mathbf{c}_{k'}^0\|^2 / \tau)
	}, 
	\quad 
	\mathbf{P}_k^0 = \sum_{s=1}^{S} 
	\alpha_k^{(s)}\,\tilde{\mathbf{Z}}^{(s)},
\end{equation}
where $\tau$ is the temperature controlling assignment sharpness. 
The resulting prototype bank 
$\mathbf{P} = [\mathbf{P}_1; \ldots; \mathbf{P}_K] \in 
\mathbb{R}^{K \times N \times D}$ is treated as a learnable parameter 
and jointly optimized with downstream tasks.

\subsubsection{Prototype-Modulated Cross-Temporal Interaction}
As illustrated in Fig.~\ref{fig:pgcai}(b), for each level $i$, 
we first construct change queries by fusing bi-temporal features 
and their difference:
\begin{equation}
	\mathbf{U}^i = \mathrm{MLP}([\mathbf{F}_1^i;\, \mathbf{F}_2^i;\, 
	|\mathbf{F}_1^i - \mathbf{F}_2^i|]) \in \mathbb{R}^{N \times D}.
\end{equation}
The query, key, and value projections are computed as:
\begin{equation}
	\mathbf{Q}^i = \mathbf{U}^i\mathbf{W}^i_Q, \quad
	\mathbf{K}^i = \mathcal{R}(\mathbf{P}\mathbf{W}^i_K), \quad
	\mathbf{V}^i = \mathcal{R}(\mathbf{P}\mathbf{W}^i_V),
\end{equation}
where $\mathcal{R}(\cdot)$ denotes the reshape operation that 
flattens the prototype bank $\mathbf{P} \in \mathbb{R}^{K \times N \times D}$ 
into $\mathbb{R}^{KN \times D}$ for attention computation. 
The modulation feature $\mathbf{M}^i$ is then retrieved via:
\begin{equation}
	\mathbf{M}^i = \mathrm{Softmax}\!\left(
	\frac{\mathbf{Q}^i(\mathbf{K}^i)^T}{\sqrt{d_k}}
	\right)\mathbf{V}^i \in \mathbb{R}^{N \times D},
\end{equation}
$\mathbf{M}^i$ encodes semantic priors about which change type each 
location belongs to, guiding bidirectional cross-temporal attention 
via element-wise modulation. The T1$\to$T2 and T2$\to$T1 directions 
are computed symmetrically:
\begin{equation}
	\mathbf{Q}^i_a = \mathbf{F}^i_a\mathbf{W}^i_{Q,a}, \quad
	\mathbf{K}^i_b = (\mathbf{F}^i_b \odot \mathbf{M}^i)\mathbf{W}^i_{K,b}, \quad
	\mathbf{V}^i_b = (\mathbf{F}^i_b \odot \mathbf{M}^i)\mathbf{W}^i_{V,b},
\end{equation}
where $\{a,b\} \in \{\{1,2\},\{2,1\}\}$ denotes the two directions 
of bidirectional attention. The output features are then obtained as:
\begin{align}
	\mathbf{G}^i_1 &= \mathrm{Softmax}\!\left(
	\frac{\mathbf{Q}^i_1(\mathbf{K}^i_2)^T}{\sqrt{d_k}}
	\right)\mathbf{V}^i_2 + \mathbf{F}^i_1, \quad
	\mathbf{G}^i_2 = \mathrm{Softmax}\!\left(
	\frac{\mathbf{Q}^i_2(\mathbf{K}^i_1)^T}{\sqrt{d_k}}
	\right)\mathbf{V}^i_1 + \mathbf{F}^i_2,
\end{align}
where $\odot$ denotes element-wise multiplication and the residual 
term preserves original features. This bidirectional block is stacked 
in two cascaded layers with independent parameters, yielding the final 
change-aware features $\{\mathbf{G}_1^i, \mathbf{G}_2^i\}_{i=1}^{4}$.

\subsection{Task-Adaptive Multi-head Gating}

Each change-aware feature $\mathbf{G}_j^i$ is decomposed into $H$ heads, 
and for each task $t \in \{d, c\}$, a head-wise sigmoid gate is applied:
\begin{equation}
	g_{j,h}^{i,t} = \sigma(\mathbf{w}_{j,h}^{i,t} \cdot 
	\mathcal{P}(\mathbf{G}_{j,h}^i) + b_{j,h}^{i,t}), 
	\quad 
	\tilde{\mathbf{G}}_{j,h}^{i,t} = g_{j,h}^{i,t} \cdot \mathbf{G}_{j,h}^i,
\end{equation}
where $\mathcal{P}(\cdot)$ denotes global average pooling. 
All levels and temporal phases are then fused via learnable weights:
\begin{equation}
	\mathbf{O}^t = \sum_{i=1}^{4}\sum_{j=1}^{2} \beta_j^{i,t}\,
	[\tilde{\mathbf{G}}_{j,1}^{i,t};\ldots;\tilde{\mathbf{G}}_{j,H}^{i,t}],
	\quad
	\beta_j^{i,t} = \frac{\exp(s_j^{i,t})}{\sum_{i',j'}\exp(s_{j'}^{i',t})}
\end{equation}
yielding task-specific representations 
$\mathbf{O}^d, \mathbf{O}^c \in \mathbb{R}^{N \times D}$.

\subsection{Task-Specific Decoders}

\subsubsection{Change Detection Decoder}
An FPN-based branch decodes $\mathbf{O}^d$ into a change probability 
map $\hat{\mathbf{M}} \in \mathbb{R}^{H \times W}$, supervised by 
binary cross-entropy:
\begin{equation}
	\mathcal{L}_d = -\frac{1}{HW}\sum_{i,j}
	\bigl[\mathbf{M}_{ij}\log\hat{\mathbf{M}}_{ij} + 
	(1-\mathbf{M}_{ij})\log(1-\hat{\mathbf{M}}_{ij})\bigr].
\end{equation}


\subsubsection{Change Captioning Decoder}
As illustrated in Fig.~\ref{fig:pgcai}(c), to inject spatial 
localization priors into caption generation, we encode 
$\mathbf{F}_{s2}$, the high-resolution feature map from the 
finest level of the FPN detection branch, which retains 
fine-grained spatial details of changed regions, into compact 
detection tokens via adaptive pooling and a two-layer MLP:
\begin{equation}
	\mathbf{D}_d = \mathrm{MLP}\bigl(
	\mathrm{Reshape}(\mathcal{A}(\mathbf{F}_{s2}, (H_m, W_m)))\bigr)
	\in \mathbb{R}^{B \times N_m \times K_d},\quad N_m = H_m W_m.
\end{equation}
Captioning features $\mathbf{O}^c$ are projected to the LM input 
space and concatenated with detection tokens:
\begin{equation}
	\mathbf{V}_{\text{comb}} = [\mathbf{D}_d;\,\mathbf{W}_c\mathbf{O}^c]
	\in \mathbb{R}^{B \times (N_m+N) \times D_{\text{LM}}}.
\end{equation}
This concatenation strategy allows the language model to 
simultaneously attend to semantic change features from 
$\mathbf{O}^c$ and explicit spatial grounding from 
$\mathbf{D}_d$, where the latter encodes the shape, extent, 
and spatial distribution of changed regions that visual 
features alone cannot precisely convey.
$\mathbf{V}_{\text{comb}}$ is fed into Qwen2-1.5B-Instruct, 
fine-tuned via LoRA to preserve pretrained linguistic knowledge 
while adapting to remote sensing change semantics. A task prompt 
instructing the model to describe observed changes based on 
change-aware and detection-derived tokens is used for 
autoregressive caption generation (see supp.).
%

\subsection{Vision-Language Semantic Alignment}
To align captioning visual representations with the CLIP semantic space, 
we apply InfoNCE contrastive loss, analogous to CLIP's image-text 
contrastive learning but using the frozen CLIP text encoder as a fixed 
semantic anchor. Text embeddings are obtained via $\mathbf{e}_t = \Phi_t(\mathbf{T})$, 
and visual embeddings are derived by pooling the LM's last-layer hidden 
states—which encode visually-grounded representations closer to natural 
language semantics than raw visual features—and projecting to the same space:

\begin{equation}
	\small
	\mathbf{e}_v^c = \mathrm{Linear}\!\left(\frac{1}{L}\sum_{l=1}^{L}
	\mathbf{h}_l^{\mathrm{LM}}\right) \in \mathbb{R}^{d_t}, \qquad
	\mathcal{L}_a = -\frac{1}{B}\sum_{b=1}^B \log
	\frac{\exp(\mathrm{sim}(\mathbf{e}_v^{c,(b)},\mathbf{e}_t^{(b)})/\tau)}
	{\sum_{b'=1}^B \exp(\mathrm{sim}(\mathbf{e}_v^{c,(b)},\mathbf{e}_t^{(b')})/\tau)}.
\end{equation}

\subsection{Training Objective and Strategy}
The overall training objective jointly optimizes three losses, 
where the captioning branch minimizes token-level cross-entropy 
against ground-truth captions:
\begin{equation}
	\mathcal{L}_{\text{tot}} = \lambda_1(t)\,\mathcal{L}_c 
	+ \lambda_2(t)\,\mathcal{L}_d + 0.3\,\mathcal{L}_a, \qquad
	\mathcal{L}_c = -\tfrac{1}{n}\sum_{i=1}^{n}\sum_{v=1}^{V}
	y_{i,v}\log y'_{i,v}.
\end{equation}
\textbf{Dynamic Weight Balancing.} Following~\cite{dwa}, 
$\lambda_1(t)$ and $\lambda_2(t)$ are adapted based on 
each task's loss improvement rate:
\begin{equation}
	\lambda_k(t) = \frac{2\exp(w_k(t{-}1)/T)}
	{\sum_{i=1}^{2}\exp(w_i(t{-}1)/T)},
	\quad
	w_k(t{-}1) = \frac{\mathcal{L}_k(t{-}1)}{\mathcal{L}_k(t{-}2)},
	\quad k\in\{1,2\},
\end{equation}
where $T{=}2.0$ controls the sharpness of weight allocation,
assigning higher weights to tasks with slower improvement.

\section{UCCD Dataset}

\subsection{Data Collection and Preprocessing}

This paper constructs a change detection and captioning 
dataset UCCD (Urban Change Captioning and Detection) oriented toward 
urban construction scenarios. Image data were collected in Xuzhou City, 
Jiangsu Province, China, using DJI drones with nadir viewing angles. 
The original resolution is 3024×4032 pixels with a spatial resolution 
of 6 cm/pixel. The temporal interval between each image pair is 
approximately 7 days, a time window that effectively captures urban 
construction dynamics. To ensure image registration accuracy, the LightGlue~\cite{lindenberger2023lightglue} algorithm is 
employed for pixel-level alignment of all image pairs. After 
registration, the original images are cropped into 1024×1024 pixel 
sub-images using a sliding window strategy, and samples in which the changed region occupies less than 0.8\% of the total image area are discarded to ensure sufficient change content. After the above processing, 9,000 
high-quality image pairs are ultimately obtained, which are split into 
training, validation, and test sets at a ratio of 7:1:2.

\subsection{Dataset Annotation}
\begin{figure*}[t]
	\centering
	\includegraphics[width=\textwidth]{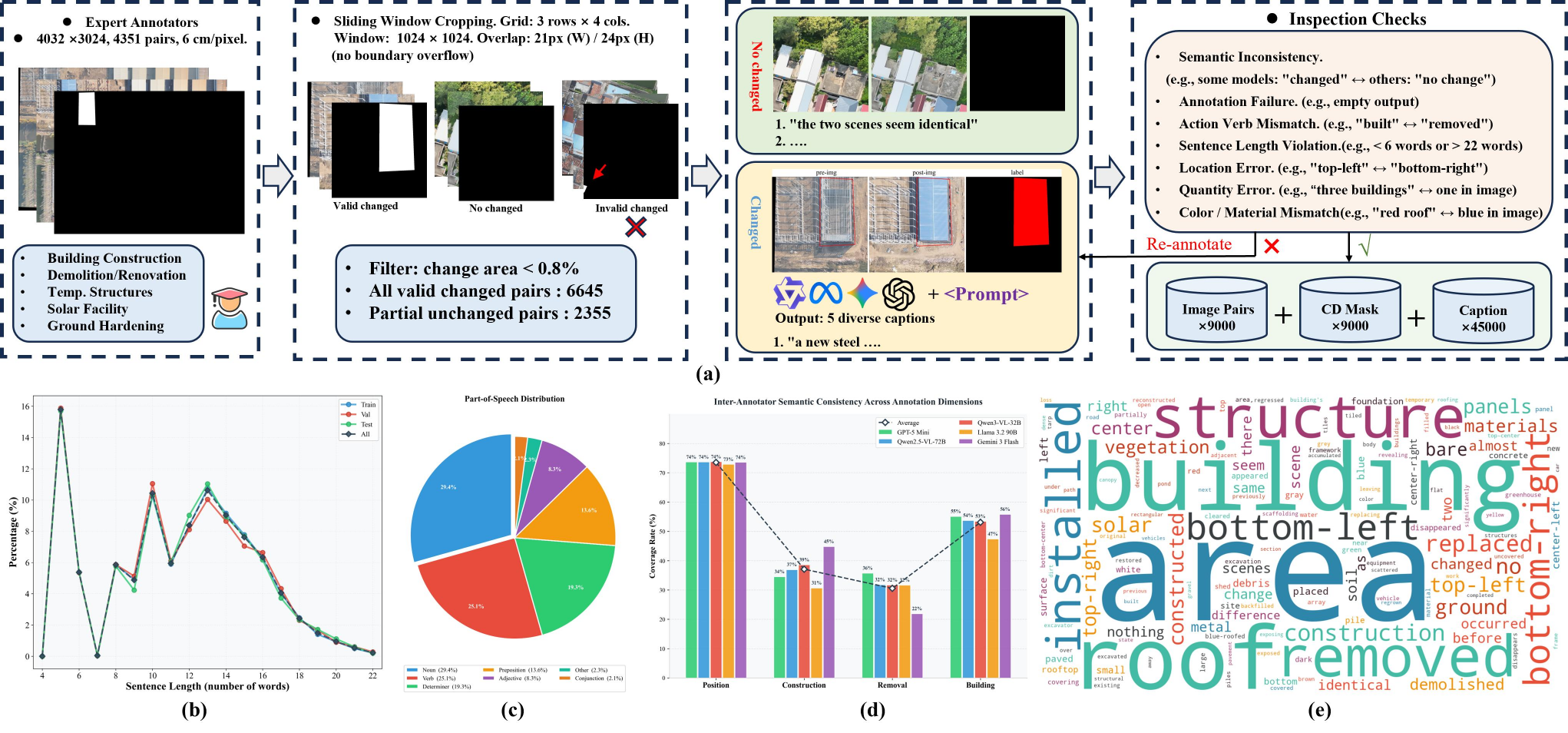}
	\caption{Overview of the UCCD dataset construction and statistical analysis.
		(a) Data annotation pipeline for UCCD dataset construction.
		(b) Sentence length distribution across Train, Val, and Test splits. 
		(c) Part-of-speech distribution of all captions, with nouns (29.4\%) 
		and verbs (25.1\%) dominating, reflecting the action-oriented nature 
		of change descriptions. 
		(d) Inter-annotator semantic consistency across four key annotation 
		dimensions (Position, Construction, Removal, and Building), 
		demonstrating high agreement among the five annotation models. 
		(e) Word cloud visualization.}
	\label{fig:dataset_analysis}
\end{figure*}

The dataset contains 6,645 changed image pairs and 2,355 
unchanged image pairs, with category distribution reflecting 
actual patterns in short-term urban monitoring. Change types cover 
multiple dimensions including building construction/demolition/
renovation, temporary structure installation, solar facility 
installation, ground hardening and land use changes, and vegetation 
removal and planting. Table~\ref{tab:dataset_comparison} compares 
UCCD with existing datasets. Figure~\ref{fig:dataset_analysis} presents 
the full annotation pipeline and statistical analysis of the dataset.

\begin{table*}[t]
	\centering
	\caption{Comparison of UCCD with existing change captioning datasets.}
	\label{tab:dataset_comparison}
	\resizebox{\textwidth}{!}{
		\renewcommand{\arraystretch}{0.5}
		\setlength{\tabcolsep}{3pt}
		\tiny
		\begin{tabular}{lcccccc}
			\toprule
			Dataset & Image Size & Resolution & Pairs & Texts & Annot. & Scene Type \\
			\midrule
			DUBAI-CCD \cite{hoxha2022change} & 50$\times$50 & 30 m & 500 & 2,500 & Manual & Urban/Suburban \\
			LEVIR-CCD \cite{hoxha2022change} & 256$\times$256 & 0.5 m & 500 & 2,500 & Manual & Urban \\
			LEVIR-CC \cite{liu2022levir} & 256$\times$256 & 0.5 m & 10,077 & 50,385 & Manual & Urban \\
			SECTION \cite{yang2025restricted} & 256$\times$256 & 0.3--3 m & 4,059 & 12,200 & Manual & Urban/Rural \\
			LEVIR-MCI \cite{liu2024changeagent} & 256$\times$256 & 0.5 m & 10,077 & 50,385 & Manual & Urban \\
			LEVIR-CDC \cite{shi2024multitask} & 256$\times$256 & 0.5 m & 10,077 & 50,385 & Manual & Urban \\
			WHU-CDC \cite{10740028} & 256$\times$256 & 0.075 m & 7,434 & 37,170 & Manual & Urban \\
			SECOND-CC \cite{karaca2025secondcc} & 256$\times$256 & 0.3--3 m & 6,041 & 30,205 & Manual & Urban/Rural \\
			\midrule
			\textbf{UCCD (Ours)} & 1024$\times$1024 & 0.06 m & 9,000 & 45,000 & Auto. & UAV Urban \\
			\bottomrule
		\end{tabular}
	}
\end{table*}

\textbf{Change Mask Annotation.} All change masks are manually 
annotated by professional annotators to ensure precise spatial 
localization.

\textbf{Text Caption Annotation.} We adopt a multi-model collaborative automated annotation strategy, 
selecting five mainstream vision-language large models as annotators: 
GPT-5 Mini, Qwen2.5-VL-72B, Qwen3-VL-32B, Llama 3.2 90B Vision, and Gemini 2.0 Flash. Each model independently generates one caption, and their 
differing inference preferences and expression styles naturally 
produce caption diversity, providing rich language supervision signals 
for model training. The prompt template provides scene metadata, 
bi-temporal image structure, change type definitions, and output 
format constraints to ensure consistent and accurate caption 
generation. For unchanged image pairs, a fixed sentence set strategy 
is adopted to avoid model hallucination, with five standard captions 
assigned to each pair: \textit{``the two scenes seem identical'', 
	``the scene is the same as before'', ``there is no difference'', 
	``no change has occurred''}, and \textit{``almost nothing has 
	changed''}. Annotations with generation failures or logical 
inconsistencies—such as mismatched location, color, or action 
descriptions—are identified through manual inspection and 
re-annotated to ensure overall quality. Upon completion of the above 
pipeline, 500 captions are randomly sampled for human evaluation, 
yielding an annotation consistency rate exceeding 98\%, which 
confirms the reliability of the annotation process. The dataset 
contains a total of 45,000 annotated sentences. More details about 
UCCD are provided in the supplementary material.

\section{Experiments}

\subsection{Evaluation Metrics and Datasets}
\textbf{Change Captioning:} BLEU-1/2/3/4 \cite{papineni-etal-2002-bleu} measure n-gram overlap; METEOR \cite{banerjee-lavie-2005-meteor} considers synonyms and stemming matches; ROUGE-L \cite{lin-2004-rouge} evaluates based on longest common subsequence; CIDEr-D \cite{vedantam2015cider} measures consistency with multiple reference texts and serves as the core metric. \textbf{Change Detection:} IoU~~\cite{zhou2019iou} and F1-score~\cite{ghasemian2024fusion} measure pixel-level overlap between predicted masks and ground truth masks. \textbf{Datasets}: \textbf{WHU-CDC} and \textbf{UCCD}.

\subsection{Implementation Details}
The proposed method trained on two NVIDIA A100 GPUs. We employ the AdamW 
optimizer with a global initial learning rate of 1e-4, while the vision 
encoder and projection layers use a learning rate of 1e-5, and weight 
decay coefficient is set to 5e-4. Training proceeds for 200 epochs with 
a total batch size of 32. We apply LoRA 
with rank $r{=}16$ and $r{=}64$ to the CLIP vision encoder and Qwen2-1.5B 
language decoder, respectively. For WHU-CDC and UCCD, the number of prototype clusters $K$ is set 
to 5 and 8, respectively, determined by the semantic diversity 
of change types in each dataset.

\subsection{Comparison with State-of-the-Art Methods}

Table~\ref{tab:combined_results} presents performance comparisons on WHU-CDC 
and UCCD. PTNet achieves state-of-the-art results on 
both benchmarks while maintaining a compact model size of 165.71M parameters.

\textbf{WHU-CDC.}
PTNet achieves the best results on all seven metrics. 
Compared to the strongest baseline KCFI, B1 through B4 improve by 
0.60, 0.62, 0.54, and 0.90, respectively, and METEOR improves by 0.74, 
reflecting more accurate semantic matching. Against Semantic-CC, 
B1 through B4 gains are 1.17, 1.57, 1.35, and 0.94. 
ROUGE-L at 79.64 and CIDEr-D at 150.02 remain competitive, 
validating caption structural quality and consistency with reference texts.

\textbf{UCCD.}
PTNet again leads all metrics, with B1--B4 at 83.26, 75.35, 70.44, 
and 66.89, surpassing KCFI by 1.08, 0.73, 0.66, and 0.95, respectively, 
and METEOR reaching 44.15. CIDEr-D reaches 188.35, exceeding KCFI by 2.67, 
benefiting from UCCD's five-caption-per-sample design. 
The generally elevated CIDEr-D scores on UCCD, for instance 
Prompt-CC achieves 184.82 compared to 134.50 on WHU-CDC, 
reflect richer reference annotations in this dataset.

\textbf{Model Efficiency.}
With only 165.71M parameters, PTNet is significantly more 
compact than KCFI at 309.55M, Semantic-CC at 299.89M, and 
Chg2Cap at 285.50M, while outperforming all of them on every 
metric, demonstrating a strong accuracy--efficiency trade-off.

\begin{figure*}[t]
	\centering
	\includegraphics[width=1\textwidth]{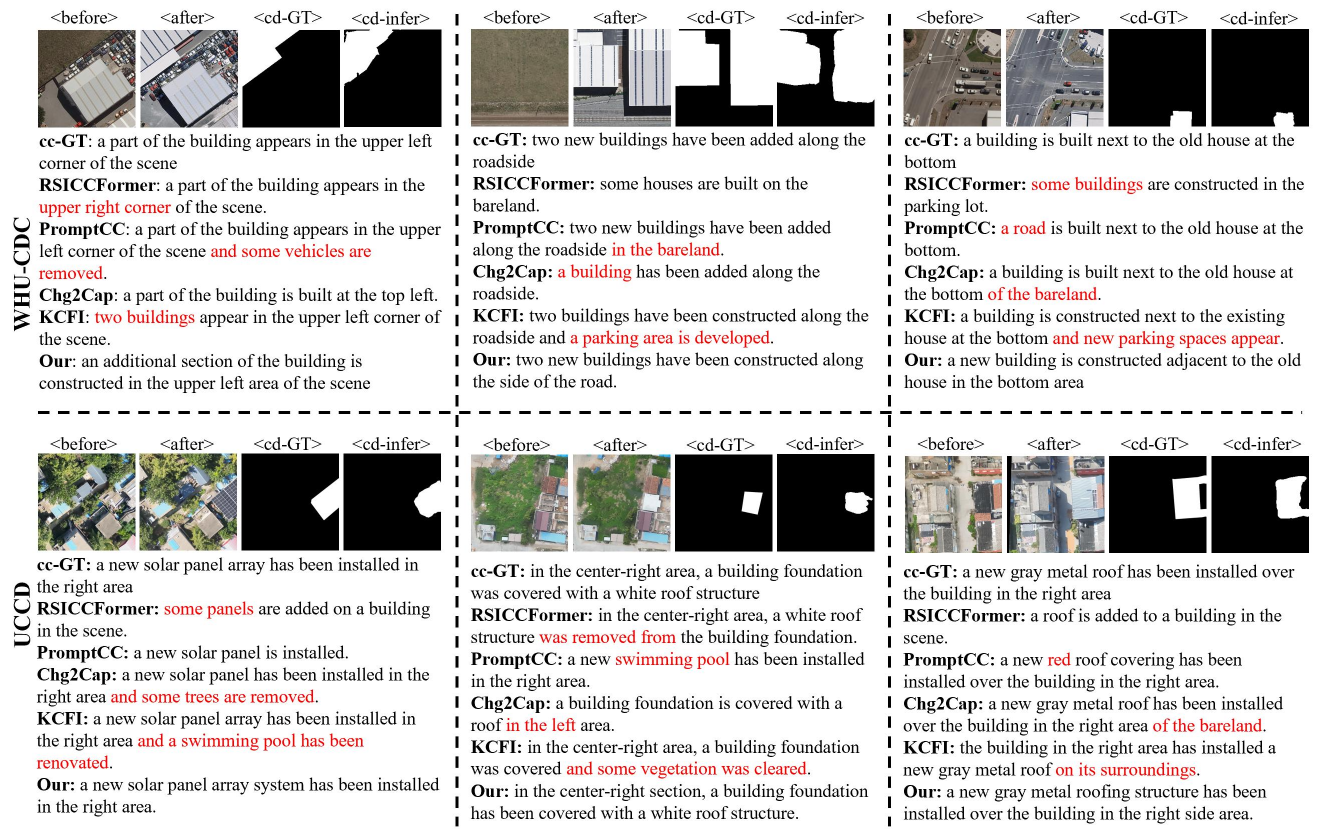}
	\caption{
		Qualitative comparison on WHU-CDC and UCCD. 
		\textcolor{red}{Red} text highlights erroneous or hallucinated 
		descriptions.
	}
	\label{fig:qualitative}
\end{figure*}

Figure~\ref{fig:qualitative} presents qualitative comparisons on both 
datasets. For each sample, we show the bi-temporal image pair, 
ground-truth mask (\texttt{cd-GT}), predicted mask (\texttt{cd-infer}), 
and captions from each method, with incorrect descriptions highlighted 
in \textcolor{red}{red}. Competing methods frequently produce hallucinated 
or spatially imprecise descriptions, such as misidentifying change locations 
or confusing change types, whereas PTNet generates more accurate 
and spatially faithful captions. This qualitative advantage stems from the 
explicit spatial priors injected by the detection branch via the mask 
encoder, and the cross-modal alignment enforced by the contrastive 
learning constraint. Notably, PTNet demonstrates robustness even in challenging 
cases such as partially occluded structures and densely 
built regions, where competing methods frequently produce 
location errors or hallucinated object descriptions. 

Although change detection is an auxiliary task, our method achieves 
the high F1 and IoU. Against the specialized 
detector SARAS-Net, F1 improves by 2.03/0.87 on UCCD/WHU-CDC, 
showing that PG-CAI provides effective semantic priors for 
detection. Against joint-training methods KCFI and Semantic-CC, 
IoU improves by up to 1.32, attributed to TAMG supplying 
finer-grained spatial features for the detection branch while 
avoiding task-feature conflicts.

\begin{table*}[t]
	\centering
	\caption{Performance comparison. B1--B4: BLEU-1--4; 
		MET: METEOR; R-L: ROUGE-L; CID: CIDEr-D. \textbf{Bold}: best result; 
		\underline{underline}: second best.}
	\label{tab:combined_results}
	\resizebox{\textwidth}{!}{%
		\begin{tabular}{lcccccccccccccccccccc}
			\toprule
			\multirow{2}{*}{Method} 
			& \multicolumn{9}{c}{WHU-CDC} 
			& \multicolumn{9}{c}{UCCD} 
			& \multirow{2}{*}{\makecell{Train \\ Params (M)}} \\
			\cmidrule(lr){2-10} \cmidrule(lr){11-19}
			& B1 & B2 & B3 & B4 & MET & R-L & CID & F1 & IoU
			& B1 & B2 & B3 & B4 & MET & R-L & CID & F1 & IoU & \\
			\midrule
			ChangeFormer~\cite{bandara2022changeformer}
			& -- & -- & -- & -- & -- & -- & -- & 86.11 & 77.88
			& -- & -- & -- & -- & -- & -- & -- & 68.03 & 50.46 & 41.03 \\
			SARAS-Net~\cite{chen2023saras}
			& -- & -- & -- & -- & -- & -- & -- & 88.90 & 83.22
			& -- & -- & -- & -- & -- & -- & -- & 70.74 & 56.26 & 56.89 \\
			\midrule
			MCCFormers-S~\cite{qiu2021mccformers}
			& 81.12 & 75.04 & 69.95 & 65.34 & 42.11 & 78.52 & 147.09 & -- & --
			& 79.85 & 72.18 & 66.32 & 61.28 & 40.33 & 76.85 & 178.42 & -- & -- & 135.01 \\
			RSICCformer~\cite{liu2022remote}
			& 80.05 & 74.24 & 69.61 & 66.54 & 42.65 & 73.91 & 133.44 & -- & --
			& 78.92 & 71.54 & 65.87 & 62.15 & 40.87 & 72.34 & 165.28 & -- & -- & 172.80 \\
			PSNet~\cite{liu2023progressive}
			& 81.26 & 73.25 & 65.78 & 60.32 & 36.97 & 71.60 & 130.52 & -- & --
			& 80.21 & 70.68 & 62.95 & 56.84 & 35.62 & 69.92 & 158.76 & -- & -- & 231.23 \\
			Prompt-CC~\cite{liu2023promptcc}
			& 81.12 & 73.96 & 67.22 & 61.45 & 36.99 & 71.88 & 134.50 & -- & --
			& 81.13 & 72.49 & 65.11 & 58.22 & 38.28 & 71.00 & 184.82 & -- & -- & 196.28 \\
			Chg2Cap~\cite{chang2023chg2cap}
			& 78.93 & 72.64 & 67.20 & 62.71 & 41.46 & 77.95 & 144.18 & -- & --
			& 77.68 & 70.28 & 64.58 & 59.84 & 39.92 & 76.28 & 176.54 & -- & -- & 285.50 \\
			DiffusionRSCC~\cite{yu2025diffusion}
			& 75.32 & 70.15 & 66.40 & 63.76 & 40.18 & 73.80 & 127.96 & -- & --
			& 74.26 & 67.82 & 63.75 & 60.94 & 38.65 & 72.14 & 155.42 & -- & -- & 58.42 \\
			Semantic-CC~\cite{liu2024semanticcc}
			& 82.77 & 76.32 & 71.59 & 68.43 & 44.49 & 78.23 & \underline{150.23} & \underline{88.46} & \underline{84.95}
			& 81.54 & 73.86 & 68.95 & 65.72 & 42.76 & 76.58 & \underline{186.47} & \underline{71.94} & \underline{56.45} & 299.89 \\
			KCFI~\cite{yang2025kcfi}
			& \underline{83.34} & \underline{77.27} & \underline{72.40} & \underline{68.47} 
			& \underline{44.95} & \underline{79.59} & 149.32 & 88.75 & 84.35
			& \underline{82.18} & \underline{74.62} & \underline{69.78} & \underline{65.94} 
			& \underline{43.28} & \underline{78.12} & 185.68 & 71.36 & 55.87 & 309.55 \\
			\midrule
			\textbf{PTNet}
			& \textbf{83.94} & \textbf{77.89} & \textbf{72.94} & \textbf{69.37} 
			& \textbf{45.69} & \textbf{79.64} & \textbf{150.02} 
			& \textbf{89.77} & \textbf{86.27}
			& \textbf{83.26} & \textbf{75.35} & \textbf{70.44} & \textbf{66.89} 
			& \textbf{44.15} & \textbf{78.47} & \textbf{188.35}
			& \textbf{72.77} & \textbf{57.65} & 165.71 \\
			\bottomrule
	\end{tabular}}
\end{table*}

\begin{table*}[t]
	\centering
	\caption{Ablation study on WHU-CDC and UCCD.}
	\label{tab:ablation}
	\resizebox{\textwidth}{!}{%
		\small
		\begin{tabular}{cccc|ccccccc|ccccccc}
			\toprule
			\multicolumn{4}{c|}{Configuration} 
			& \multicolumn{7}{c|}{WHU-CDC} 
			& \multicolumn{7}{c}{UCCD} \\
			\cmidrule(lr){1-4} \cmidrule(lr){5-11} \cmidrule(lr){12-18}
			Proto & TAMG & Det. Guided & $\mathcal{L}_a$ 
			& B-1 & B-2 & B-3 & B-4 & M & R-L & C
			& B-1 & B-2 & B-3 & B-4 & M & R-L & C \\
			\midrule
			-- & -- & -- & --
			& 79.87 & 73.62 & 68.37 & 64.83 & 42.53 & 76.48 & 137.64
			& 79.18 & 71.27 & 65.73 & 62.31 & 41.08 & 74.87 & 173.18 \\
			\checkmark & -- & -- & --
			& 81.43 & 75.31 & 70.28 & 66.68 & 43.76 & 77.73 & 142.57
			& 80.79 & 72.83 & 67.58 & 64.08 & 42.29 & 76.28 & 179.06 \\
			\checkmark & \checkmark & -- & --
			& 82.68 & 76.58 & 71.63 & 67.93 & 44.68 & 78.61 & 146.13
			& 81.97 & 74.03 & 68.87 & 65.41 & 43.17 & 77.31 & 183.47 \\
			\checkmark & \checkmark & \checkmark & --
			& 83.47 & 77.43 & 72.47 & 68.84 & 45.31 & 79.28 & 148.63
			& 82.81 & 74.87 & 69.83 & 66.34 & 43.79 & 78.07 & 186.72 \\
			\checkmark & \checkmark & \checkmark & \checkmark
			& \textbf{83.94} & \textbf{77.89} & \textbf{72.94} & \textbf{69.37} 
			& \textbf{45.69} & \textbf{79.64} & \textbf{150.02}
			& \textbf{83.26} & \textbf{75.35} & \textbf{70.44} & \textbf{66.89} 
			& \textbf{44.15} & \textbf{78.47} & \textbf{188.35} \\
			\midrule
			
	\end{tabular}}
\end{table*}

\begin{table*}[t]
	\centering
	\caption{Left: Ablation on detection branch components 
		(\checkmark/--: enabled/disabled). 
		Right: Ablation on prototype modulation and initialization in PG-CAI.}
	\label{tab:ablation_combined}
	\begin{minipage}[t]{0.54\textwidth}
		\centering
		\resizebox{\linewidth}{!}{%
			\scriptsize
			\begin{tabular}{ccc|cccc|cccc}
				\toprule
				\multicolumn{3}{c|}{Configuration}
				& \multicolumn{4}{c|}{WHU-CDC}
				& \multicolumn{4}{c}{UCCD} \\
				\cmidrule(lr){1-3}\cmidrule(lr){4-7}\cmidrule(lr){8-11}
				Det. Dec. & TAMG & Det. G.
				& B-4 & M & R-L & C
				& B-4 & M & R-L & C \\
				\midrule
				-- & -- & --
				& 67.83 & 44.12 & 78.34 & 146.28
				& 65.28 & 42.83 & 77.12 & 184.23 \\
				\checkmark & -- & --
				& 68.04 & 43.88 & 78.01 & 147.43
				& 65.67 & 42.59 & 76.88 & 185.47 \\
				\checkmark & \checkmark & --
				& 69.02 & 44.87 & 79.18 & 149.12
				& 66.38 & 43.64 & 77.98 & 187.12 \\
				\checkmark & \checkmark & \checkmark
				& \textbf{69.37} & \textbf{45.69}
				& \textbf{79.64} & \textbf{150.02}
				& \textbf{66.89} & \textbf{44.15}
				& \textbf{78.47} & \textbf{188.35} \\
				\bottomrule
		\end{tabular}}
	\end{minipage}
	\hfill
	\begin{minipage}[t]{0.43\textwidth}
		\centering
		\resizebox{\linewidth}{!}{%
			\scriptsize
			\begin{tabular}{lcc|cccc|cccc}
				\toprule
				\multirow{2}{*}{Configuration}
				& \multirow{2}{*}{Mod.}
				& \multirow{2}{*}{Init}
				& \multicolumn{4}{c|}{WHU-CDC}
				& \multicolumn{4}{c}{UCCD} \\
				\cmidrule(lr){4-7}\cmidrule(lr){8-11}
				& & & B-4 & M & R-L & C
				& B-4 & M & R-L & C \\
				\midrule
				w/o Modulation
				& -- & --
				& 67.85 & 44.21 & 78.65 & 146.83
				& 65.32 & 43.18 & 77.45 & 185.12 \\
				Random Init
				& \checkmark & Rand.
				& 68.43 & 44.38 & 78.89 & 146.52
				& 65.67 & 43.25 & 77.63 & 184.87 \\
				Proto. Init
				& \checkmark & Proto.
				& \textbf{69.37} & \textbf{45.69}
				& \textbf{79.64} & \textbf{150.02}
				& \textbf{66.89} & \textbf{44.15}
				& \textbf{78.47} & \textbf{188.35} \\
				\bottomrule
		\end{tabular}}
	\end{minipage}
\end{table*}

\subsection{Ablation Studies}
\subsubsection{Component-wise Ablation}
We ablate each component, 
starting from a baseline using bi-temporal feature 
concatenation with shared representation.

\textbf{Proto.}
Prototype-modulated cross-temporal attention yields the gain (B-4: +1.85/+1.77, CIDEr-D: 
+4.93/+5.88 on WHU-CDC/UCCD). Unlike simple feature 
subtraction, which treats all spatial positions equally, 
change-type prototypes provide structured semantic priors 
that steer cross-temporal attention toward semantically 
coherent regions, enabling more discriminative temporal 
correspondences.

\textbf{TAMG.}
Task-adaptive gating further improves B-4 by 1.25/1.33. 
Without it, detection and captioning share identical 
representations, creating a granularity conflict between 
spatial precision and semantic abstraction. Head-level 
sigmoid gates disentangle these demands within a unified 
feature space, benefiting both tasks simultaneously.

\textbf{Det. Guided.}
Injecting detection-derived spatial priors into the 
captioning branch yields consistent gains (B-4: 
+0.91/+0.93), as detection tokens explicitly supply 
\emph{where} and \emph{how extensively} change occurred—
information that visual features alone cannot precisely 
convey.

\textbf{$\mathcal{L}_a$.}
The contrastive alignment loss improves METEOR by 
+0.38/+0.36 and CIDEr-D by +1.39/+1.63. The pronounced 
METEOR gain reflects InfoNCE's nature: aligning visual 
representations to CLIP text embeddings cultivates 
synonym-aware semantics, which METEOR directly rewards.

Overall, PTNet surpasses the baseline by 4.54/4.58 in B-4 
and 12.38/15.17 in CIDEr-D on WHU-CDC/UCCD. The larger 
CIDEr-D gain on UCCD stems from its five-caption-per-sample 
design, which amplifies the benefit of cross-modal alignment.

\begin{table*}[t]
	\centering
	\caption{Performance comparison.}
	\label{tab:main}
	\resizebox{\textwidth}{!}{%
		\begin{tabular}{lcccccccccccccc}
			\toprule
			\multirow{2}{*}{Method} & \multicolumn{6}{c}{WHU-CDC} & \multicolumn{6}{c}{UCCD} & \multirow{2}{*}{Params(M)} \\
			\cmidrule(lr){2-7} \cmidrule(lr){8-13}
			& B-4 & MET & R-L & CID & F1 & IoU & B-4 & MET & R-L & CID & F1 & IoU & \\
			\midrule
			Chg2Cap       & 62.71 & 41.46 & 77.95 & 144.18 & --    & --    & 59.84 & 39.92 & 76.28 & 176.54 & --    & --    & 285.5 \\
			Prompt-CC     & 61.45 & 36.99 & 71.88 & 134.50 & --    & --    & 58.22 & 38.28 & 71.00 & 184.82 & --    & --    & 196.3 \\
			Semantic-CC   & 68.43 & 44.49 & 78.23 & 150.23 & 88.46 & 84.95 & 65.72 & 42.76 & 76.58 & 186.47 & 71.94 & 56.45 & 299.9 \\
			KCFI          & 68.47 & 44.95 & 79.59 & 149.32 & 88.75 & 84.35 & 65.94 & 43.28 & 78.12 & 185.68 & 71.36 & 55.87 & 309.5 \\
			\textbf{PTNet (Ours)} & \textbf{69.37} & \textbf{45.69} & \textbf{79.64} & \textbf{150.02} & \textbf{89.77} & \textbf{86.27} & \textbf{66.89} & \textbf{44.15} & \textbf{78.47} & \textbf{188.35} & \textbf{72.77} & \textbf{57.65} & \textbf{165.7} \\
			\bottomrule
	\end{tabular}}
\end{table*}
\subsubsection{Analysis of Detection Branch}
Table~\ref{tab:ablation_combined} (left) ablates each component 
of the detection branch. Adding the detection decoder alone yields 
marginal captioning gains but slightly degrades METEOR, indicating 
feature competition between the two tasks without proper decoupling. 
TAMG resolves this conflict at the attention-head level, producing 
the largest improvement among the three components. Further 
incorporating detection-derived spatial priors via Det. Guided 
yields consistent gains across all metrics on both datasets, 
completing PTNet.

\subsubsection{Analysis of Prototype Initialization in PG-CAI}
Table~\ref{tab:ablation_combined} (right) investigates prototype 
modulation and initialization strategy in PG-CAI.
Removing prototype modulation causes consistent drops across 
all metrics (CIDEr-D: 150.02→146.83 on WHU-CDC), confirming 
that semantic modulation is essential for cross-temporal 
interaction. Random initialization reveals a telling 
dichotomy: n-gram metrics (B-4, ROUGE-L) slightly improve 
over the no-modulation baseline, yet semantic metrics 
(METEOR, CIDEr-D) degrade. This suggests random prototypes 
introduce spurious biases that improve lexical overlap while 
disrupting semantic coherence—indicating that prototype 
\emph{quality} matters as much as their presence. 
Prototype-based initialization yields consistent gains 
across all metrics (CIDEr-D: +3.50/+3.48 on WHU-CDC/UCCD), 
confirming that semantically meaningful prototypes provide 
reliable modulation signals.

\section{Conclusion}
In this work, we advance the problem of urban change detection 
and captioning from both data and methodology perspectives. 
On the data side, we present UCCD, the first large-scale 
UAV-based benchmark designed for urban construction scenarios, 
covering diverse change types with high-resolution imagery 
and multi-model collaborative annotations, providing the 
community with a valuable resource beyond existing 
satellite-based benchmarks. On the methodology side, we 
propose PTNet, which explicitly models change-type semantics 
via a learnable prototype bank, decouples the feature 
requirements of detection and captioning through task-adaptive 
gating, and bridges the two tasks by injecting detection-derived 
spatial priors into the caption generation process. Extensive 
experiments on WHU-CDC and UCCD demonstrate that PTNet achieves 
state-of-the-art performance on both tasks across all metrics. 
We hope this work inspires broader interest in fine-grained 
urban change understanding, moving beyond simple binary change 
masks toward spatially-aware and semantically-rich change 
descriptions.


%
%
\renewcommand{\bibsection}{\section*{references}}
\bibliographystyle{unsrt}
\bibliography{main}

\end{document}